# Self-organising Urban Traffic control on micro-level using Reinforcement Learning and Agent-based Modelling

## Stefan Bosse

**University of Bremen, Dept. Mathematics & Computer Science, Bremen, Germany**

**Abstract**. Most traffic flow control algorithms address switching cycle adaptation of traffic signals and lights. This work addresses traffic flow optimisation by self-organising micro-level control combining Reinforcement Learning and rule-based agents for action selection performing long-range navigation in urban environments. I.e., vehicles represented by agents adapt their decision making for re-routing based on local environmental sensors. Agent-based modelling and simulation is used to study emergence effects on urban city traffic flows. An unified agent programming model enables simulation and distributed data processing with possible incorporation of crowd sensing tasks used as an additional sensor data base. Results from an agent-based simulation of an artificial urban area show that the deployment of micro-level vehicle navigation control just by learned individual decision making and re-routing based on local environmental sensors can increase the efficiency of mobility in terms of path length and travelling time.

## 1. Introduction

Traffic jams and disturbance in traffic flows are ubiquitous in modern cities. These phenomena effect societies and causes economic losses of relevant order. Adaptive traffic optimisation, addressing individual and domestic traffic, is crucial for improving living quality in growing cities (i.e., it is basically a scaling problem).

Traffic is a distributed complex problem with hard predictable dynamics on global and spatial scale. Arising of jams, slow down of average ensemble speed, and ensemble dead locks in traffic flows without a clearly identifiable cause are prominent examples.

A traffic situation consists of a large set of individual entities (treated as agents) that interact with each other and satisfying constraints (i.e., streets, traffic signs, traffic rules, dangerous situations, and so on). Individual traffic entities are controlled by individual humans or by machine algorithms (automatic and autonomous vehicles) based on a a set of behaviour rules. These behaviour rules can significantly influenced by

Stefan Bosse http://orcid.org/0000-0002-8774-6141

varying parameter sets (i.e., different classes of drivers and individualism of behaviour and goals).

Commonly, traffic flow is controlled via traffic signals (traffic lights) and dynamic signs (e.g., speed and street routing control) based on accumulated flow data in real-time. There are different basically spatial domains considered by controllers and learner instances that have to be distinguished: Global and urban city scale, "glocal" scale with transition from global to local spatial domains, i.e., connected groups of streets and street areas, the local scale, i.e., one street, part of a street, a crossroad and street junctions, crowds, and finally micro scale level, i.e., single vehicles or people.

Adaptive traffic flow control on different spatial and domain levels is attractive to reduce travelling times and energy consumption (i.e., air pollution), and to enable better scaling of growing city populations finally improving urban living quality and social prosperity. It is basically a distributed optimising problem of a large-scale dynamic system including chaotic effects. Traffic can be understood as the behaviour of a set of ensembles consisting of interacting bust mostly autonomous entities solving a set of constraints (e.g., road maps). Most work in this field focuses on traffic signal control (e.g., an overview can be found in [1]). Other aspects like individual driver decision making and path routing influencing traffic flows of is not considered. Although there are traffic simulation that consider driver behaviour, an assumption of average behaviour is made without considering real-world variations [2]. Only few work is known that incorporates crowd sensing data (one example can be found in [3]). Machine learning can be used to improve traffic flow control and user experience on macro- and micro-scale level (local optimisation). But the required training of ML models cannot be performed in real-world environments.

Simulations can overcome this limitation and can be used to (pre-)train machine learners and to investigate different traffic flow control and machine learning algorithms. But simulation relies commonly on simplified, averaged and unified behaviour models, simplified environments and situations. Agent-based modelling and simulation (ABMS) is suitable for large-scale, distributed, and complex dynamic systems with local interaction models [4]. Considering traffic and traffic control, agents are autonomous entities satisfying constraints.

To improve simulation results and to increase the robustness of control models to individual entity variations, the simulation can be extended by injecting digital twins of real entities (vehicles, humans). The behaviour model of the digital twins are derived from surveys performed via agent-based crowd sensing (CWS) and delivering sensor data suitable to compute a model parameter set (see [4] for details of augmented virtuality approaches).

Using CWS for traffic prediction was already evaluated as a valuable method by Wan et al. [5], but mainly distinguishing Vehicle to Infrastructure (V2I) and Vehicle to Vehicle (V2V) communications. The multi-agent system (MAS) architecture and framework enables the seamless coupling of MAS modelling, simulation, and distributed comput-

Stefan Bosse http://orcid.org/0000-0002-8774-6141

ing outside the simulation environment (e.g., performing mobile crowd sensing).

Traffic flow control should be achieved on three levels:

1. Ensemble control by the environment (using traffic signals and signs) using common traffic control algorithms based on sensor data (collected, e.g., by street cameras);

2. Individual control by driving entities. e.g., influencing routing and decision making of individuals via social media or navigation systems;

3. Local automatic group control (i.e., car-to-car communication and control) using local interaction agents.

In [6] and [7], self-organising traffic control was applied to traffic light signal switching. In this work, there is a focus on self-organising traffic flow control on individual level (level 2) by support decision making processes of drivers (or automatic or more advanced of autonomous vehicles), particular addressing short- and long-range routing.

In previous work. the influence of behaviour model variations from an average behaviour of traffic entities (drivers, passengers) were investigated by using ABMS with digital twins derived from CWS [4]. Commonly, agent-based simulation and agent-based distributed computing (ABC) performing the traffic control are separated. The approach from [4] unites simulation and real-world data processing by an unified mobile agent-model covering ABM, ABS, and ABC, enabling the tight coupling of real and virtual (simulation) worlds in real-time.

This work addresses therefore two paradigms to create smart traffic control:

1. Cooperating and interacting multi-agent systems;

2. Reinforcement learning (RL);

3. Self-organisation and self-adaptivity.

Due to hard predictable short-range interaction between traffic entities (traffic lights and signals, drivers, vehicles), behaviour variations (e.g., drivers not respecting constraints like speed limits) and the effect on global system behaviour, a model-free agent-based reinforcement learning approach is used and evaluated in this work to address self-organising traffic flow control on micro-scale entity level [1]. Self-organising is in this case implicitly performed by solving or rewarding constraints between physical entities and sensor feedback, e.g., distances between vehicles, spatial street constraints, and so on.

RL is closely related to the agent model. In [8], multi-agent systems perform distributed traffic signal control. Among distributed learning (that can be a challenge to implement it and to achieve stable convergence), distributed learning-agents can be deployed, each operating on a local state and optimising a sub-set set or one particular target variable. This approach requires co-ordination to optimise on global level implementing distributed co-ordination of exploration and exploitation (DCEE, introduced by Brys et al. [11]).

Stefan Bosse http://orcid.org/0000-0002-8774-6141

The main objective of this work is to find a multi-agent-based and self-organising urban traffic control architecture suitable to optimise traffic flow, i.e., increasing the average traffic flow speed, minimising or completely avoiding jams, minimising the travelling times with respect to passengers, and minimising energy for mobility. In contrast to other work ([1], [8], [9], [10], [11]) controlling traffic lights and signals only, this work will focus on the control of decision making processes of vehicles, drivers, and passengers only (e.g., influencing routing) incorporating experience and history situations. For the first time, conventional static traffic signal switching is assumed.

Furthermore, in contrast to major simulation work, the simulation in this work does not use a static entity behaviour (mobility) model. Instead, a parameterisable interaction and mobility model is used represented by agents. The simulation starts with agents posing an average parameter set extended with agents posing parameter set variants obtained by crowd sensing. The crowd sensing (that can be performed in real-time at simulation time) aims to create digital twins of traffic entities. A traffic entity is a part of a set of multiple classes consisting of vehicles (individual and domestic mobility), drivers, passengers, and traffic lights and signals.

The hypothesis to be tested in this work is the possibility to improve traffic flows by individual control of traffic participants with the goal to influence individual decision making processes like routing by using RL, even in the absence of adaptive TSC.

A novel hybrid agent architecture based on an coupled reactive rule-based and learning-based action selection is introduced.

## 2. Agent-based Modelling and Simulation of Traffic

This section summarises the unified agent model and simulation framework used in this work for studying traffic flows and traffic management. Details can be found in [4]. Finally, an extended agent architecture coupling the original activity and rule-based agent model with RL.s

### 2.1 Computational and Physical Agents

There are two classes of agents covered by one unified agent model that is used in this work:

1. **Physical behavioural agents** representing physical entities in virtual worlds (simulation) like vehicles or individual artificial humans;

2. **Computational agents** representing mobile software in real and virtual world, i.e., used for distributed data processing and digital communication, and used for implementing chat bots;

Stefan Bosse http://orcid.org/0000-0002-8774-6141

Both types of agents are used in the simulation, but only computational agents can migrate between the simulation world and real world environments. The computational agents are required for seamless integration of mobile crowd sensing into the simulation (optionally in real-time), discussed in the next sub-section. The agents are programmed in *AgentJS*, which is syntactilly generic *JavaScript* with some semantic modifications.

Details of the unified agent model can be found in [12], and details about the used agent processing platform *JAM* can be found in [4].

The agent behaviour model is purely reactive and state-based, shown in Fig. 1. An agent consists of code and private data (body variables). The code describes the agent behaviour consisting of activities executing actions. There are conditional and unconditional transitions between activities. The conditions access agent body variables only. Both code and data are mobile and an agent process snapshot is capable to migrate between two agent platforms. Activities of an agent represent intentions and micro goals, e.g., changing the spatial position, modifying the environment, communicating with other agents, and agent replication. Agents processes support the concept of blocking, i.e., the agent processing can be suspended during waiting for an event or the satisfaction of a constraint condition.

In contrast to commonly used reactive agent behaviour models, the ATG can be modified by the agent itself offering self-adaptivity. An agent can remove or add activities and transitions, either providing sub-classing (specialisation) or learning.

Agents can communicate via tuple spaces (data driven) or by using signal messages (agent driven). Physical and computational agents can communicate via signals to synchronise or to exchange data.

Stefan Bosse http://orcid.org/0000-0002-8774-6141

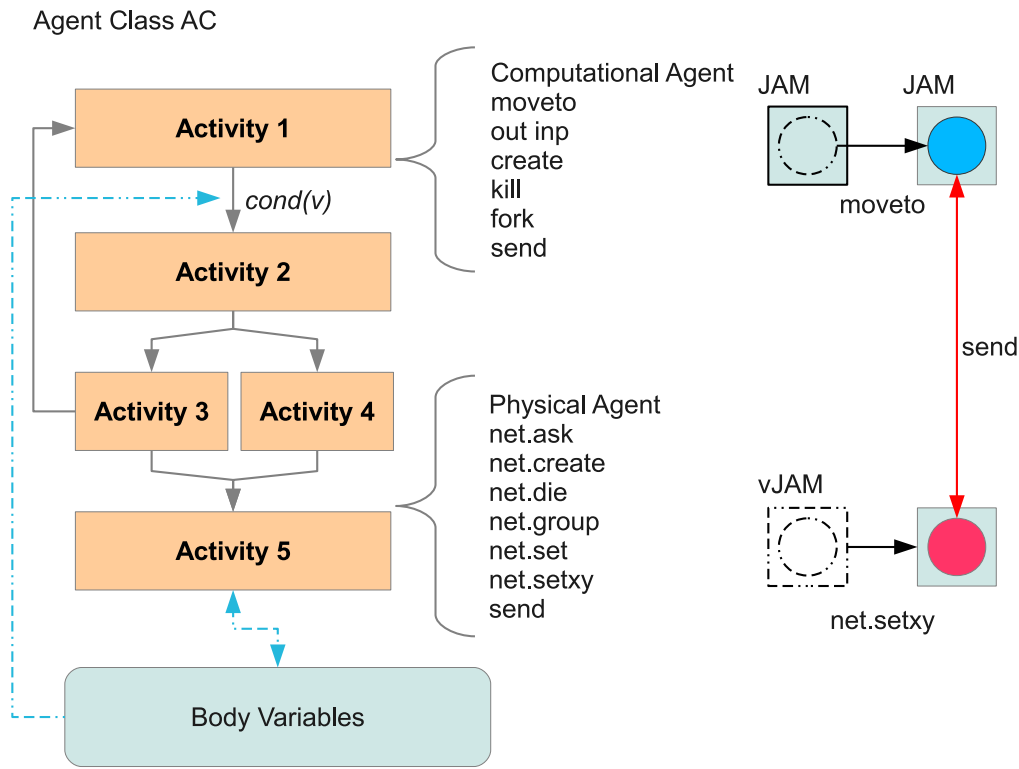

*Figure 1: Activity-Transition Graph (ATG) behaviour and data model of an agent for a specific class AC.* Physical and computational agents differ in their action set (right side).

### 2.2 Mobile Crowd Sensing

Among well generated and computed synthetic sensor data, real world data acquired from sensors is required to perform accurate simulations, too.

Commonly, simulation is performed with artificial agent models derived from theoretical considerations or experimental data. Augmented virtuality enables dynamic simulations with agents representing real entities (devices, vehicles, or crowds). By using crowd sensing it is possible to create digital twins of real entities based on a parameteriseable behaviour and interaction model (discussed later). The parameters of artificial entities in the simulation represented by physical agents are collected by sensor data, i.e., surveys optionally fusioned with physical sensors like GPS. The proposed self-organising traffic control is relying on crowd sensing. Crowd sensing can

Stefan Bosse http://orcid.org/0000-0002-8774-6141

happen between machines (Machine-Machine Interaction, MMI) and between humans and machines (HMI).

The crowd sensing via mobile agents (chat bots in case of HMI) enable the interaction of real world entities with mobile computational agents and digital twins derived from crowd sensing in the simulation world in real-time and vice versa. The digital twins as well as the artificial physical agents in the simulation can interact via communication agents and in the case of HMI by dynamically created (influenced) chat dialogues reflecting the state of the simulation world.

**2.3  Simulation Framework SEJAM2**

The framework couples virtual and real worlds by integrating simulations with human interactions by using computational agents (chat bots) and physical behavioural agents inside the simulation.

The entire simulation architecture coupling real- and virtual worlds consists of the following components, shown in Fig. 2:

1. Unified Agent Processing Platform based on JavaScript: JavaScript Agent Machine (JAM) with two architecture sub-classes

   - Physical Platform
   - Virtual Platforms (of a physical platform)

2. Crowd Sensing Software (Mobile App, WEB Browser using JAM, or Embedded Computer using JAM);

3. Agent-based Simulation on top of JAM with Internet connectivity supporting two different agent types:

   - **Physical behavioural agents**
   - **Computational agents**

4. Chat dialouges, Chat bots and Mobile Agents collecting user and device sensor data;

Virtual JAM nodes are used in the simulation to implement physical entities like vehicles, traffic control entities, or artificial humans. Physical and computational agents can interact and communicate with each other either by using the computational method of tuple-spaces and signals or in the case of physical agents only a shared-world and shared-memory method with a dedicated (NetLogo compatible) API.

Stefan Bosse http://orcid.org/0000-0002-8774-6141

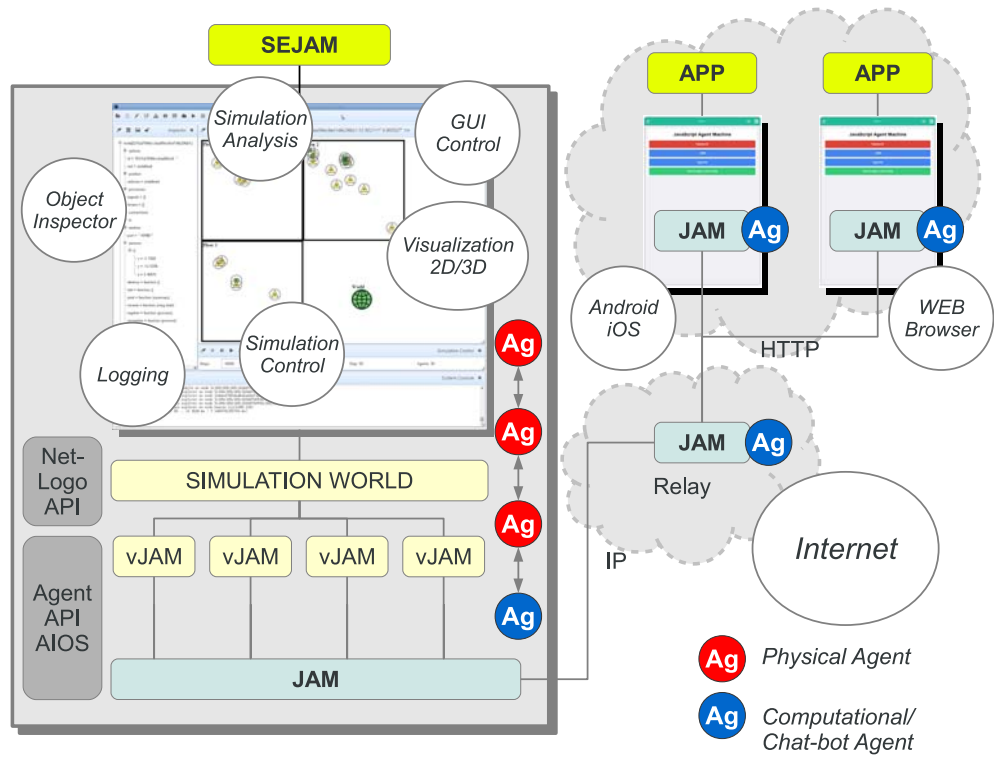

*Figure 2: Principle concept of closed-loop simulation for augmented virtuality: (Left) Simulation framework based on the JAM platform (Right) Mobile and non-mobile devices executing the JAM platform connected with the virtual simulation world (via the Internet) [4]*

The mixed-model simulation world consists of physical and computational agents bound to logical (virtual) platforms (host of the agent) that are arranged or located on a lattice to provide world discretisation (although, spatial positions in the simulation world are not restricted to a grid world). The agents are mobile. Computational agents as mobile software processes can migrate between platforms (both in virtual and real digital worlds), whereas physical agents are fixed to their platform and only the platform is mobile (virtual world).

The agents are programmed in JavaScript executed by JAM that can be deployed on a wide range of host platforms (mobile devices, servers, IoT devices, WEB browser [4]). JAM provides virtualisation by the Agent Input-Output System (AIOS), tuple spaces, and virtual (logical) nodes bound to a world contained in one physical JAM node. Each physical or logical JAM node can be connected with an unlimited number of remote

Stefan Bosse http://orcid.org/0000-0002-8774-6141

JAM nodes by physical links (UDP/TCP/HTTP using the AMP protocol). Logical nodes can be connected by virtual links. Links provide agent process migration, signal (message) and tuple propagation.

The JAM platform provides a set of ML algorithms including different RL algorithms. JAM agents are mobile, i.e., they can migrate between logical and physical JAM nodes by transferring the state and code of the agent. The platform splits learned models from algorithms enabling mobile models (e.g., decision trees or neural networks). Additionally, these models can be shared by different agents or be inherited.

### 2.4 Parameterised Agent Model

Physical agents representing traffic entities are modelled as reactive state-based agents with the following parameterisable behaviour model (with parameter set *Par*):

$$\begin{aligned} percept &: Sen \times Per \to Per \\ next &: St \times Per \times Par \times R \times C \to St \\ action &: St \times Par \to Act \end{aligned} \quad (1)$$

The perception function *percept* maps sensor input on perception states, optionally parameterised (e.g., by defining weight parameters). The parameters have impact on the state transition function *next* and the selection of appropriate actions from a set of actions *Act* by the action function *action*. For example, a vehicle controller agent supports a set of actions *Act*:

- Keep preferred direction
- Change direction (turn)
- Follow current route
- Change vehicle speed
- Change distance to front vehicle
- Start/stop driving
- Overtake

The internal state set *St* represents activities of the agent and therefore is a composition of the control and internal data state of the agent. Parameters from the set can select sub-sets of actions to adapt to specific situations. The computation of the next agent state (and partially the action selection) bases basically on a set of rules *R* and constraints *C*.

The behaviour model of JAM agents is an activity-transition graph (ATG) with activities performing actions (representing the *action* function and sub-goals of the agent) and (conditional) transitions between activities representing state transitions and the *next* function.

### 2.5 Hybrid Rule- and Learning-based Agent Architecture

Although RL agents itself pose a well defined architecture, in this work the RL instance is used as a co-function for the state transition and action computation, shown in Fig. 3. Now there are two output functions *action* and *RL* that select appropriate actions to be executed by a controller agent (i.e., controlling the vehicle). Although both output functions can select actions from the same set of actions *Act*, it is feasible to split the action set in two sub-sets $Act_1$ and $Act_2$ used by the two output functions, respectively. A final fusion of both action selections (that can be contrary or cross-forbidden) is performed by the *fusion* function.

There are three different RL algorithms mapping state variables on actions available for the agent:

- Temporal Difference Learning
- Dynamic Programming
- Deep Q Learning

The addition of the *RL* function extends the functional agent model of Eq. 1:

$$\begin{aligned}
percept &: Sen \times Per \to Per \\
next &: St \times Per \times Par \times R \times C \to St \\
action &: St \times Par \to Act_1 \\
rl &: r \times Per \times \to Act_2 \\
reward &: Act_2 \times Per \times Par \to r[-1, 1] \\
fusion &: Act_1 \times Act_2 \to Act
\end{aligned} \quad (2)$$

The parameter set can be changed at run-time by the agent itself to adapt to specific known or new situations, eventually modified by the RL instance, too. The parameters define the superposition of rule-based and learning-based action selection (in the range from 0-100%) enabling switching between both architecture and behaviour models.

The fusion function is basically a lazy constraint solver that checks the two actions provided by the rule-based action function and the predicted action from the RL function. The constraint solver checks for contradiction and invalid actions. That means it refuses invalid actions with respect to the current agent state. For instance, changing the direction is currently not possible (spatial constraints) or re-routing exceeds a specific frequency. The selected action output of the *fusion* is feed back to the *reward* function (comparing the fusioned and the predicted action).

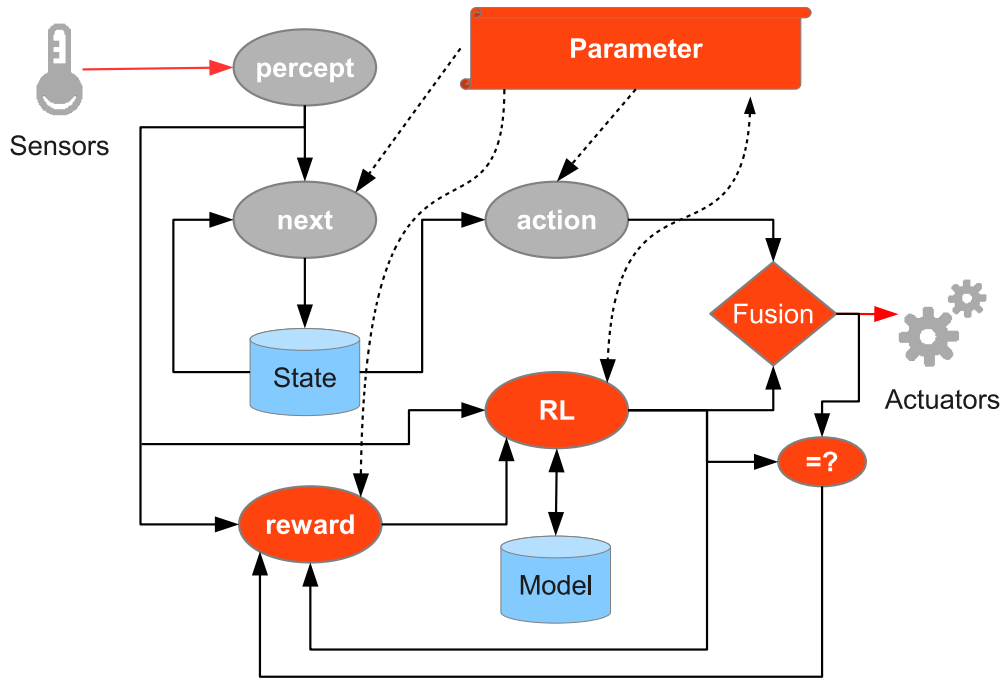

*Figure 3. The proposed hybrid parameterised agent architecture combining reactive state-based action selection with RL*

## 3. Coupling Multi-Agent Systems and Reinforcement Learning for Traffic control

The self-organised traffic control (in sense of optimisation) MAS consists of the following agent classes:

- Local vehicle controller agents performing basic rule-based car control and short-range navigation
- Communication agents (bridging vehicles and TSC entities)
- Navigation agents (coupled to vehicles) performing adaptive and optimised long-range navigation using Reinforcement learning (RL)
- Global and local traffic controller agents (rather simple in this work)

The simulation adds the following agents:

- Entity agent (e.g., driver, vehicle, ..) representing an artificial entity;
- Digital Twin agent representing a real entity; and a
- Simulation controller agent (world agent).

Reinforcement learning is commonly modelled as a Markov decision process (MDP) and is used to learn and optimise sequential decision making processes, typical in traffic and navigation. An RL instance maps environmental states on actions that are performed by a physical or computational agent. RL can be classified in model-based and model-free learning (e.g., Q-learning, applied to traffic control [9]). Only the latter class is considered in this work because it is commonly not possible or meaningful to derive interaction models in complex traffic scenarios.

RL can be applied to traffic signal control (RL-TSC) on global or local environmental domain level and/or to entity control on individual level, e.g., vehicle control. Hierarchical and domain-based learning was proposed by Abdoos et al. [10]. An RL instance is associated to a learner agent that can be coupled to other agents, like traffic control or driver agents. The RL agent outputs a recommendation for actions to be considered by a controller agent, e.g., speed limitation, traffic light switching, or route planning and decision making. There are centralised Single Agent RL (SARL) and decentralised Multi Agent RL (MARL) approaches. This work covers a hierarchical MARL approach.

RL requires sensor input $s$ and a feedback via utility and conflict functions defining the reward function $u(s)$. An RL model outputs an action $a$ from a set of possible actions $A$ executed by a controller agent. The sensor system for traffic lights is usually based on stationary vehicle detectors, like inductive loops or cameras using vision algorithms to identify vehicle flows. Most traffic control algorithms consider only actual traffic situations without considering predictions of future variations, flows, and context changes. The sampling of vehicle data is often inaccurate, e.g., the speed of vehicles, introducing sources of error. Other variables influencing traffic like crowd flows, events (emergencies, work closings, shopping), and time-specific crowd flow variations are commonly not considered in traffic flow control. Sensory input from social media and other urban sensors can be considered by ML, too.

The main issue in RL-TSC is the determination of the state of the environment from sensors that is represented to the RL agent. Typical state variables (on a macro-scale level) are:

- Queue length $ql$
- Waiting time $qt$
- Flow rate $fr$
- Averaged vehicle speed (normalised to speed limit) $va$
- Signal change times and delays

- Energy (electrical energy or fuel consumption)
- User satisfaction (overall utility) of all traffic participants (including pedestrians)

Commonly, RL agents act autonomously and individually. RL agents can interact via communication, but there is no initial coupling of the learning instances themselves. Another approach for coupling RL agents is a hierarchical organisation of the learners shown in Fig. 4. There are different domains ranging from a macro-level (city, urban area, spatially in the kilo meter range) to a micro-level (streets, crossroads, and individual entities with interaction in the meter range).

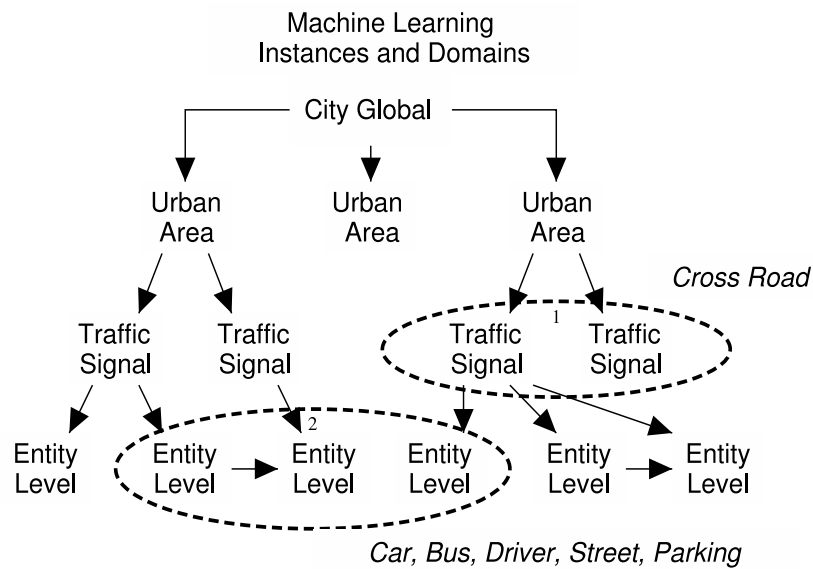

Figure 4. Hierarchical decomposition of machine learning instances for urban traffic control

The environmental domains are covered by different learning instances. In case of hybrid learning systems, the output of lower learning instance levels can be used as sensors input for higher levels and vice versa.

With respect to the micro-scale level, further state variables are introduced to provide sensors $S$ of variables capable to detect traffic situations from the view of point of single vehicles:

- Normalised average speed $v_0$
- Distance to front and back neighbour vehicles $df_0$, $db_0$

- Distance to destination *de*, progress $\Delta de_0$
- Direction to destination $td_0$
- Direction of vehicle $r_0$
- Queuing time $qt_0$
- Possible driving and turning directions (with respect to street constraints) in left, right, and backward directions *tl, tr, tb*
- A set of possible paths from current position to destination *P*

These state variables are primarily used to adapt the vehicle driving control based on self-organisation of local ensembles and to recognise jam situations (present and future situations). For each state variable $x_0$ there is a desired value $x_1$. Fig. 5 shows the principle MAS configuration, the sensors of the agents, and their communication paths. Vehicle agents (consisting of a coupled controller and learner agent pair) can sense their neighbourhood using their own sensors, the sensors from neighbouring vehicles, and the sensors of nearby traffic light signals stations.

The traffic agent uses the following sensor variables:

- Queuing length *ql*
- Queuing time *qT*
- Average vehicle flow speed *vf*
- Flow rate *fr*

Stefan Bosse http://orcid.org/0000-0002-8774-6141

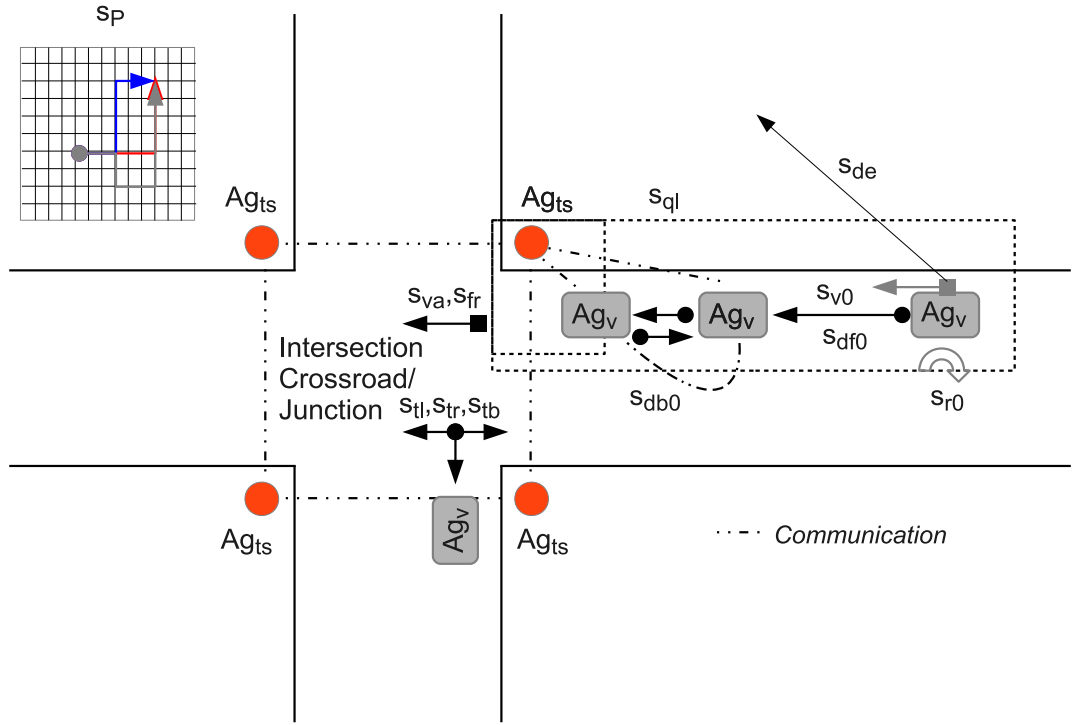

*Figure 5. Agents, their sensors ($s_x$ with respect to the sensor variable x), and agent communication paths ($Ag_{ts}$: Traffic sign agent, $Ag_v$: Vehicle agent)*

The reward function returning a value in the range [-(a+b+c),(a+b+c)] (a,b,c are weight factors) of the vehicle agent uses the superposition of the ration of actual and average speed, actual and average queuing time, and the progress to reach the destination:

$$r = a\frac{v_0 - \overline{v_0}}{max(v_0, \overline{v_0})} + b\frac{qt_0 - \overline{qt_0}}{max(qt_0, \overline{qt_0})} + c\frac{sd - \overline{sd}}{max(sd, \overline{sd})} \qquad (3)$$

Vehicle agents can communicate with neighbouring nodes to get group (vehicle ensemble) sensor data (e.g., queue length). Additionally, vehicle agents nearby a traffic light signal can communicate with the traffic controller agent to get switching information (remaining time of green/red phase, switching times, ..) and traffic flow sensor data.

As learned RL models are portable and mobile, they can be shared (copied) by a set of agents selecting the best trained models from a set of models. Since RL is an incremental learning method, the pre-trained models are improved during run-time (applica-

Stefan Bosse http://orcid.org/0000-0002-8774-6141

tion and learning) with new feedback (reward).

## 4. Agent Behaviour

To summarise the vehicle controller agent, as long as a vehicle is moving it recalculates the speed, direction, and turns for re-routing the path from a source to a destination location. There is a set of rules and activities enabling driving fusioned with a set of actions to change routing based on learning and sensor data to improve the reward (i.e., satisfying individual goals).

The more general agent behaviour introduced in the previous section is now refined and simplified to reduce the degree of freedom in the action space and to reduce the state space.

**4.1 Vehicle Agent**

The vehicle agent is responsible to implement decision tree and rule-based automatic driving to satisfy the following constraints:

1. Driving on the right side of a street in the currently selected direction (North, South, West, East)

2. The vehicle speed $v$ may not be higher than the speed limit on the current road: $v < v_{max}$

3. The distance to the next vehicle ahead may not be lower than a speed-dependent distance limit: $df < df_{min}$

4. There may no collision (two vehicle may not occupy the same place): $(x_a, y_a) \neq (x_b, y_b)$

Any violation of the above constraints results in the execution of an action of the set of actions $A$:

1. Moving one step left, right, or ahead: $|\Delta|=1$;

2. increasing or decreasing the vehicle speed;

3. stopping movement.

A rule-based decision tree is used to select an action $a$ of the set of actions $A$;

### 4.2 Navigation Agent

The navigation agent is responsible for long-range navigation to optimise the following target variables (odometry measures):

1. Path distance AB

2. Mean velocity for path AB

3. Mean travelling time for path AB

The odometry measures are normalised to the never reachable shortest and fastest path routing possible without any traffic control and only one vehicle moving on a street.

## 5. Experimental Results by Simulation

The experiments were performed with the SEJAM2 simulation framework. The simulation world (shown in Fig. 6) consists of an artificial street map with 14 long streets (36 street segments), 49 crossroads and junctions, and 144 traffic light signals. For the sake of simplicity the world is discretised by a mesh grid (100 × 100 cell patches). the simulation tool is not limited to discrete simulations world. The dark grey fields between streets are occupied by buildings and parking areas. The entire street area is segmented in 3330 patches. A crossing can be occupied by up to 9 vehicles. Each vehicle occupies one patch field. Up to 12 vehicles can occupy a street segment between crossings in each direction, resulting in a global maximal capacity of 2016 vehicles (although, any mobility is inhibited with this maximal population). Simulations show an increase of a jam probability starting already with 15% of street coverage!

The simulation was carried out with 100, 200 and 300 vehicles carrying a vehicle control and navigation twin agent with a static parameter set. The vehicle agent represents a physical vehicle and performs rule-based short-range navigation (collision avoidance, track following, trap/jam escape control).

A driver navigation agent (twin) performing rule- and learning-based long-range navigation was assigned to each vehicle agent composing a mobile agent group.

For a first evaluation of the new micro-level traffic control approach, fixed green-red cycles (50%-50% duty cycle) and mutual exclusive switching of perpendicular crossings of streets are assumed.

Stefan Bosse http://orcid.org/0000-0002-8774-6141

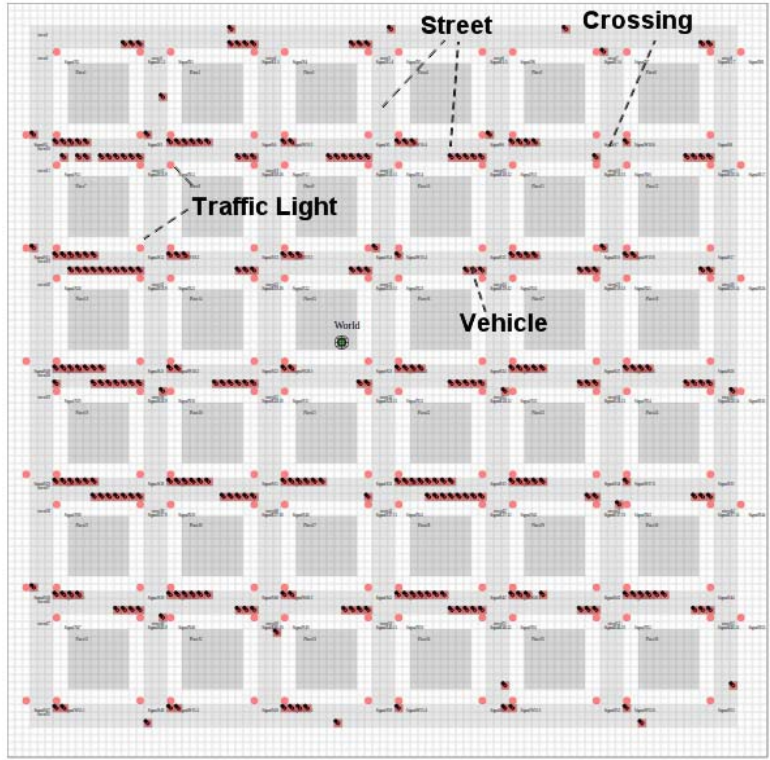

*Figure 6. The simulation world consists of streets, street segments, traffic light signals, crossroads (junctions), and vehicles occupying street patches*

Each vehicle agent has a randomly chosen start and end point. After a vehicle agent reaches its destination it restarts driving to its original starting point and vice versa. After some start-up time the simulation has a constant average number of vehicles.

A simplified RL function was chosen with a sub-set of vehicle state variables and a sub-set of vehicle control actions:

$$rl(tl, tr, tb, de, df, db, qt, ql, r) : (tl, tr, tb, de, df, db, qt, ql, r) \rightarrow \\ \{left, right, backward, speed+, speed-\} \quad (4)$$

All three available RL algorithms (TBD, DP, DQNN) were available and compared. The Deep Q-Learning algorithm with a neural network suggesting actions to be performed is the only suitable algorithm since it relies on a non-finite state world. It showed good convergence but with high learning time requirements. The DQN algorithm is used in the following simulations only.

Stefan Bosse http://orcid.org/0000-0002-8774-6141

Some examples of simulation results are shown in Fig. 7 and Fig. 8 with training runs of navigation agents. The simulation starts with a fixed population of vehicles placed randomly on street locations. The RL instances of each vehicle are initialised with random neural network weights.

The learning is performed on-line by the navigation agents with an increasing influence on the route planning (short-scale and long-scale). The training phase was performed with vehicles driving between a start and end point (randomly chosen). Each time a vehicle reaches (the randomly chosen) destination the source and destination points are exchanged and the navigation starts again.

A first approach used randomly initialised (i.e., untrained) RL networks (models). First navigation attempts are purely random walk. A second approach uses pre-training that was performed (with four vehicles) selecting a set of the best trained navigators finally passed to the post-training phase with a large number of vehicles.

The average vehicle speed was 1/5 world grid steps / simulation step and about 5000 training events (rewarding actions) were applied to each the RL instance. The entire simulation run consist of 1 Million simulation steps (real running time about 300 CPU computation minutes).

The path efficiency η is the mean ratio of the actual routed path length and the shortest possible path. The time efficiency τ is the mean ratio of actual path travelling time and the shortest possible time (assuming the shortest path and the default vehicle speed). The mean navigation error accumulates wrong predictions of a direction change by the navigation RL instance. Not possible direction changes are filtered by an higher rule-based level using vehicle sensors and constraints.

The second plot shows the accumulated mean average reward of all navigation RL instances (starting with negative values at the beginning of the simulation that is not shown in the plot). Again, the navigation RL models were not pre-trained and therefore there is an expected increase of the navigation error in first learning phase up to 10k simulation steps.

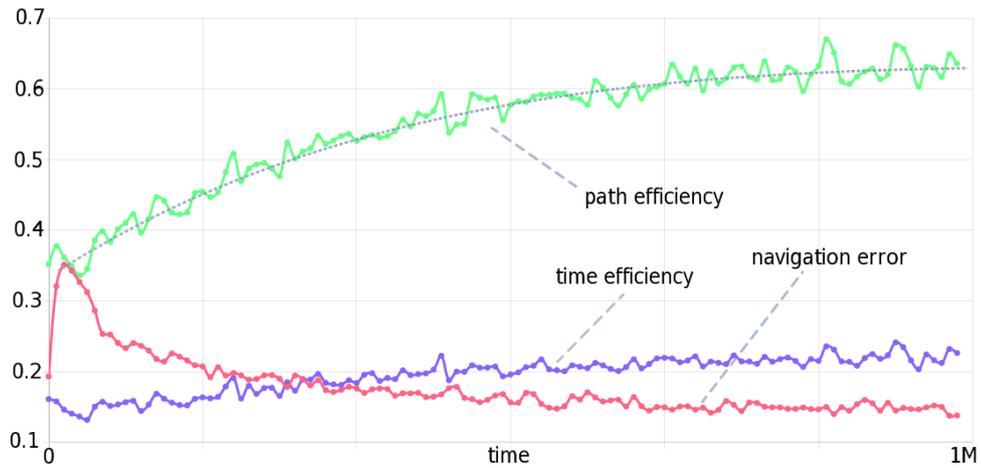

*Figure 7. Progress of navigation path efficiency, navigation error fraction, and travelling time efficiency over learning time (simulation steps, with 200 vehicles, total occupied street capacity 10%, no pre-trained RL networks)*

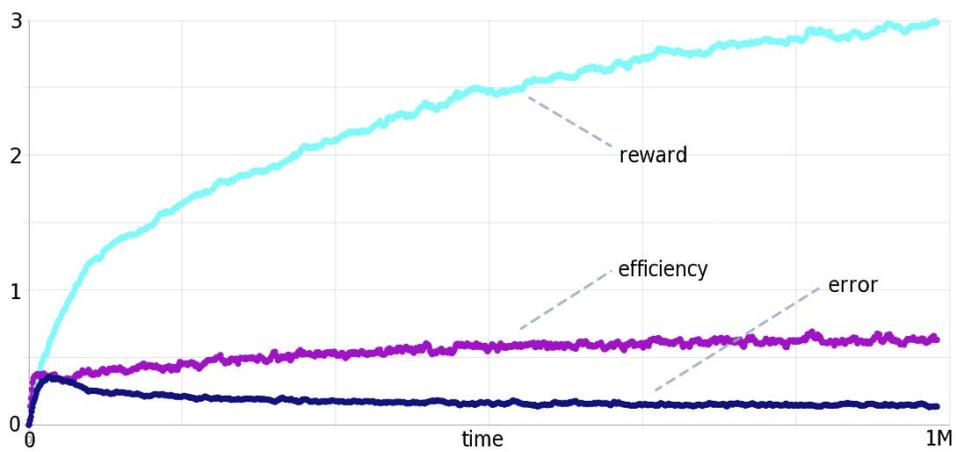

*Figure 8. Progress of global accumulated learning reward (again comparing with progress of navigation error prediction fraction and navigation efficiency) over learning time (simulation steps, with 200 vehicles, total occupied street capacity 10%, no pre-trained RL networks)*

Stefan Bosse http://orcid.org/0000-0002-8774-6141

Figures 9 and 10 show simulations results with the half of vehicles (100). There is no significant difference compared with the 200 vehicle simulation.

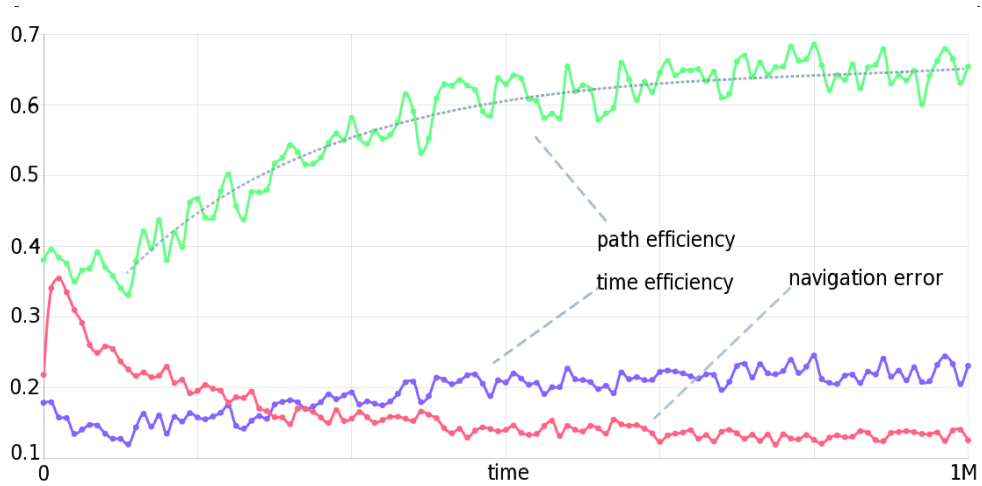

*Figure 9. Progress of navigation path efficiency, navigation error fraction, and travelling time efficiency over learning time (simulation steps, with 100 vehicles, total occupied street capacity 5%, no pre-trained RL networks)*

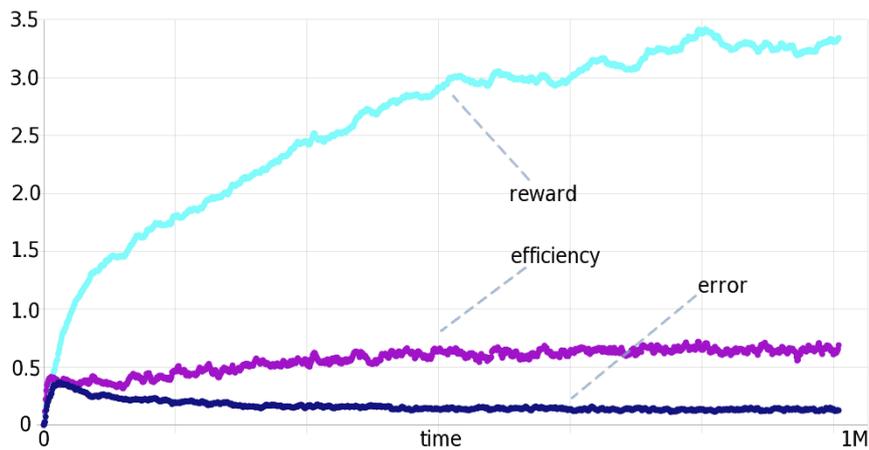

*Figure 10. Progress of global accumulated learning reward (again comparing with progress of navigation error prediction fraction and navigation efficiency) over learn-*

*ing time (simulation steps, with 100 vehicles, total occupied street capacity 5%, no pre-trained RL networks)*

The simulation with 300 vehicles starting from scratch (without any pre-trained models) results in large clusters and jams (mostly in the corner of the city world) with a high probability within the first 50k simulation steps, shown in Figure 11. Although, the automatic low-level short range navigation control contains trap escaping algorithms a highly populated area with restricted escape choices is not able to overcome this jam trap situation (just by missing choices to change the current position). Blocking situations and jams result in a significant increase of computation time slowing down the simulation significantly by an order of 10.

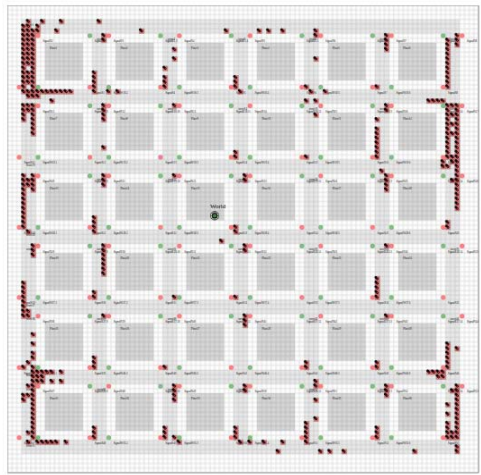

*Figure 11. Clustering and jams after 10k simulation steps (300 vehicles, total occupied street capacity 15%, no pre-trained RL networks)*

But by using pre-trained RL navigation models there are no clustering or jams observed. Each navigation agent of a vehicle uses a pre-trained model randomly selected from a set of four models. The results are shown in Figure 12 and 13. The averaged global travelling path efficiency reaches a value about 0.5 quickly. There is only a small progress by post-training (applied again continuously during simulation) with respect to path efficiency and reward. The results from Figure 7 and 9 ($\eta > 0.6$) will not be reached. This can be a result of missing model variance (only a set of four pre-trained models were provided for all 300 navigation agents following totally different paths). The global average reward is also lower ($r = 2$) compared with the not pre-trained system.

Stefan Bosse http://orcid.org/0000-0002-8774-6141

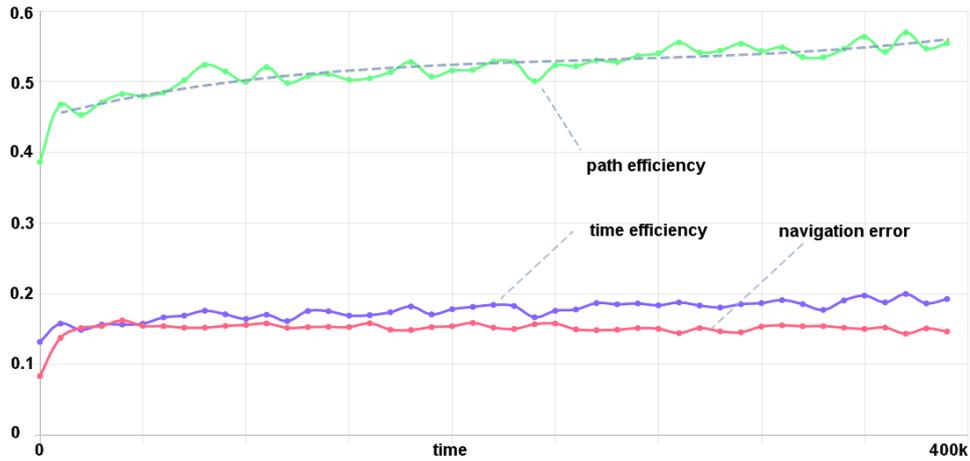

*Figure 12. Progress of navigation path efficiency, navigation error fraction, and travelling time efficiency over learning time (simulation steps, with 300 vehicles, total occupied street capacity 15%, with pre-trained RL networks)*

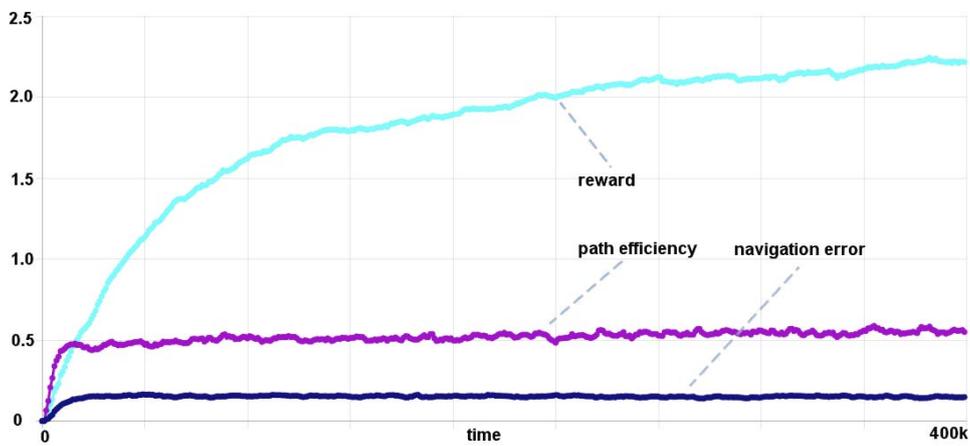

*Figure 13. Progress of global accumulated learning reward (again comparing with progress of navigation error prediction fraction and navigation efficiency) over learning time (simulation steps, with 300 vehicles, total occupied street capacity 15%, with pre-trained RL networks)*

*23*

To conclude: The deployment of self-organising micro-level vehicle control just by individual decision making and re-routing based on local environmental sensors can be used to implement self-adaptive and self-organising route navigation to avoid jams and slowdowns in traffic and to increase the global travelling efficiency in terms of route/path lengths and travelling times.

## 6. Conclusion

In contrast to common traffic management controlling traffic lights and signals only, this work addressed traffic flow optimisation on micro-level by adapting decision making processes of vehicles, primarily long-range navigation and re-routing, optionally with vehicle speed control.

In a simulation vehicles were represented by vehicle agents provided with an extended set of sensors. The behaviour model of agents is an activity-transition graph (ATG) with activities performing actions (representing the *action* function and sub-goals of the agent) and (conditional) transitions between activities representing state transitions and the *next* function. The agent model is related to the classical reactive state-based agent model. Rule-based action selection was extended by an hybrid approach with RL and reward functions. The learning navigation agents can be used directly in real-world vehicles since the agent processing platform used in the simulation is usable in technical systems, too. The agent behaviour has only be modified slightly.

Training of reinforcement learning navigation agents by thousands of trial-and-error cycles requires a long time to reach a satisfying navigation strategy better than random walk and is only possible in simulation worlds. Otherwise domestic traffic would collapse if performed in real world.

Simulation results from an agent-based simulation of an artificial urban area show that the deployment of such a micro-level vehicle control just by individual decision making, learning, and re-routing based on local environmental sensors can reach near optimal routing still under high traffic densities (regarding total route length and travelling times).

Further investigations have to be carried out to evaluate the global emergence and stability. The decision making of vehicle agents relies on rules and a black-box function learned from only a few state variables. One highly interesting aspect to be considered and evaluated is the possibility to transfer already learned models to other vehicle agents introducing multi-agent co-operation.

## 7. References

Stefan Bosse http://orcid.org/0000-0002-8774-6141